# In-situ and Non-contact Etch Depth Prediction in Plasma Etching via Machine Learning (ANN & BNN) and Digital Image Colorimetry

*Minji Kang[1,2,†], Seongho Kim[1,2,†], Eunseo Go[3], Donghyeon Paek[1], Geon Lim[4], Muyoung Kim[1], Soyeun Kim[5], Sung Kyu Jang[6], Min Sup Choi[2], Woo Seok Kang[1,7], Jaehyun Kim[8], Jaekwang Kim[8,\*], Hyeong-U Kim[1,9,\*]*

[1]Semiconductor Manufacturing Research Center, Korea Institute of Machinery and Materials (KIMM), Daejeon 34103, Republic of Korea

[2]Department of Materials Science and Engineering, Chungnam National University (CNU), Daejeon 34134, Republic of Korea

[3]Department of Organic Materials Engineering, Chungnam National University (CNU), Daejeon 34134, Republic of Korea

[4]Department of Laser & Electron Beam Technologies Research, Korea Institute of Machinery and Materials (KIMM), Daejeon 34103, Republic of Korea

[5]Department of Physic and Astronomy, DGIST, Daegu 42988, Republic of Korea

[6]Electronic Convergence Material and Device Research Center, Korea Electronics Technology Institute (KETI), Seongnam 13509, Republic of Korea

[7]Mechanical Engineering, KIMM Campus, University of Science & Technology (UST), Daejeon 34113, Republic of Korea

[8]Department of Mechanical and Design Engineering, Hongik University, Sejong 30016, Republic of Korea

[9]Nano-Mechatronics, KIMM Campus, University of Science & Technology (UST), Daejeon 34113, Republic of Korea

[†]These authors contributed equally to this work.

E-mail: guddn418@kimm.re.kr, jk12@hongik.ac.kr

# Abstract

**Abstract**

Precise monitoring of etch depth and the thickness of insulating materials, such as Silicon dioxide and silicon nitride, is critical to ensuring device performance and yield in semiconductor manufacturing. While conventional ex-situ analysis methods are accurate, they are constrained by time delays and contamination risks. To address these limitations, this study proposes a non-contact, in-situ etch depth prediction framework based on machine learning (ML) techniques. Two scenarios are explored. In the first scenario, an artificial neural network (ANN) is trained to predict average etch depth from process parameters, achieving a significantly lower mean squared error (MSE) compared to a linear baseline model. The approach is then extended to incorporate variability from repeated measurements using a Bayesian Neural Network (BNN) to capture both aleatoric and epistemic uncertainty. Coverage analysis confirms the BNN's capability to provide reliable uncertainty estimates. In the second scenario, we demonstrate the feasibility of using RGB data from digital image colorimetry (DIC) as input for etch depth prediction, achieving strong performance even in the absence of explicit process parameters. These results suggest that the integration of DIC and ML offers a viable, cost-effective alternative for real-time, in-situ, and non-invasive monitoring in plasma etching processes, contributing to enhanced process stability, and manufacturing efficiency.

## 1. Introduction

Silicon dioxide ($SiO_2$) and silicon nitride ($SiN_x$) are essential insulating materials in the fabrication of major semiconductor devices, including metal-oxide-semiconductor field-effect transistors (MOSFETs),[1-3] dynamic random-access memory (DRAM),[4] and NAND flash memory,[5] owing to their excellent dielectric properties, chemical stability, and process compatibility. The thickness of these insulating layers critically influences the device performance and reliability. In MOSFETs, the gate oxide thickness determines the capacitance and threshold voltage,[4, 6] and in DRAMs, dielectric thickness deviations impact data retention and leakage current characteristics.[7] Furthermore, non-uniformity in etch stop layer thickness during etching can result in over-etching or undercutting, leading to structural damage and degraded device yield.[8, 9] Precise in-situ monitoring and control of insulator thickness during processing are therefore indispensable for ensuring both electrical performance and manufacturing consistency.

However, most etch process evaluations are performed via ex-situ analysis after the wafers are transferred from the chamber following process completion.[10-13] This approach inevitably delays fabrication due to wafer transport and exposes the wafer to atmospheric conditions, increasing the risk of surface contamination. Moreover, disturbing the vacuum environment inside the chamber can compromise process reproducibility.[14, 15] Although in-situ ellipsometer has been widely used monitoring etch depth during processing, it typically requires relatively long measurement times and presents challenges for integration into pre-installed equipment.

In recent years, digital image colorimetry (DIC) has emerged as a promising alternative for addressing these limitations. DIC analyzes color by digitizing images acquired via image acquisition tools such as mobile phones, digital cameras, webcams, and scanners.[16] Especially, DIC has attracted attention in research fields that require non-contact, and real-time analysis. It has been widely applied in the fields of food sciences,[17-21] chemistry,[22-24] environmental sciences,[25] and biosciences,[26, 27] offering a more convenient and cost-effective alternative to traditional methods without the need for expensive equipment. The integration of DIC with machine learning (ML) algorithms has been shown to enhance precision, enable automation, and support real-time predictions. Moreover, applying ML to DIC overcomes the limitations of conventional direct color-matching approach by enabling data-driven predictions.[28-30]

In this study, a non-contact etch depth prediction method is proposed using red, green, and blue (RGB) data extracted via DIC, along with key process parameters. Etch depth, measured by ellipsometer, is employed as the output variable, and an Artificial Neural Network (ANN) is used to model its relationship with the collected input data. In plasma etching processes, acquiring nanometer-scale etch depth measurements is inherently challenging due to the limitations of in-situ measurement and the limited available data samples.[31] To address these challenges, we also incorporate a Bayesian Neural Network (BNN) framework,[32] which models the weights as probability distributions to account for both data variability and model uncertainty, thereby enabling the estimation of prediction reliability.[33] This study demonstrates that non-contact, DIC-based data can be effectively used not only for etch depth prediction but also for broader applications in process monitoring through various ML techniques. The proposed approach offers significant potential for in-situ monitoring in semiconductor manufacturing, contributing to improved process stability and yield through non-invasive observation of etching conditions.

## 2. Experimental details

In this study, ICP type reactive ion etcher (RIE; RAINBOW4420, Lam Research, Fremont, CA, USA) was used for plasma etching experiments. A high-density plasma was generated by applying radio frequency (RF) generator at 13.56 MHz to the top copper coil, While an additional RF generator was connected to the bottom electrode to regulate the DC bias voltage. Before the etching, the chamber was evacuated to a base pressure of $10^{-6}$ Torr using a turbomolecular pump (BOC Edwards, Burgess Hill, UK) to ensure a clean and stable environment. A pre-cleaning process was carried out to remove residual contaminants by introducing a gas mixture of 50 sccm Ar and 20 sccm $O_2$, while applying 250 W to the top coil for 5 minutes.

Following the cleaning process, a coupon wafer (1.5 × 1.5 $cm^2$, 300 nm $SiO_2$) was loaded into the chamber and the etching process was performed using a gas mixture of $CF_4$, $O_2$, and Ar. The flow rates of $O_2$ and Ar were fixed at 10 sccm, while the $CF_4$ flow rate was varied across four conditions, 5, 10, 15, and 20 sccm. The etching time was fixed at 180 seconds, and chamber pressure was varied across three levels: 20, 30, and 40 mTorr. The top power was varied from 50 to 110 W in 10 W increments, while the bottom power was maintained at 20 W. These experimental conditions resulted in a total of 84 conditions, summarized in **Table 1.** During the etching process, plasma characteristics were monitored in real-time using optical emission spectroscopy (OES; Maya2000Pro, Ocean Insight, Orlando, FL, USA). The OES system captures the intensity and wavelength variations of plasma emission, which reflect the interaction of reactants and by-products species during the etching process.[34, 35] Spectral data were transmitted through an optical fiber positioned at the center of the view window.

**Table 1** Process condition of ICP-RIE for SiO$_2$ etching.

| Parameter | Unit | Conditions |
|---|---|---|
| Chamber pressure | mTorr | 20, 30, 40 |
| Plasma power (Top) | W (watt) | 50, 60, 70, 80, 90, 100, 110 |
| Plasma power (Bottom) | W (watt) | 20 |
| Process time | sec | 180 |
| Gas (CF$_4$) | sccm | 5, 10, 15, 20 |
| Gas (Ar) | sccm | 10 |
| Gas (O$_2$) | sccm | 10 |

The etched SiO$_2$ samples were characterized using a spectroscopic ellipsometer (M-2000V, J.A. Woollam, Lincoln, Nebraska, USA) to confirm the thickness of SiO$_2$. Measurements were performed over and energy range of 1.0 to 3.5 eV, corresponding to wavelengths from 354 to 1240 nm. In addition, images for color analysis were captured using a smartphone camera (iPhone 12, Apple, Cupertino, CA, USA). To ensure consistent ambient lighting and photographic conditions, the coupon wafers were placed on a white background during image acquisition. Color data were extracted from a representative rectangular region on each sample. The red (R), green (G), blue (B) color ($0 \leq R, G, B \leq 255$) were quantified using the Image J software. **Figure 1** presents a schematic of the etch depth prediction framework, illustrating the overall workflow of data acquisition and model training. In this approach, in-situ RGB data obtained from DIC is paired with ellipsometer measurements of etch depth. (Additionally, in-situ OES data is utilized to verify plasma stability) These datasets are used to train ANN and BNN models for the purpose of predicting etch depth based on real-time, non-contact measurement.

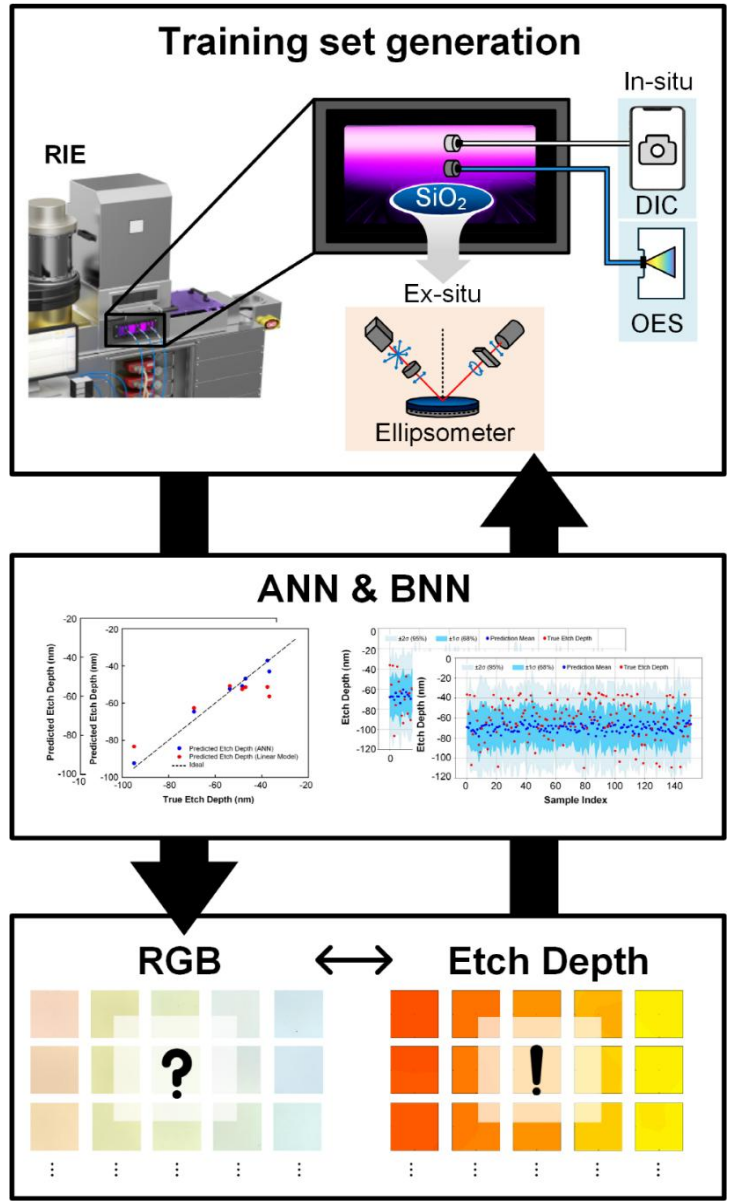

**Figure 1** Schematic illustration of the etch depth prediction framework using in-situ measurements and ML. During the RIE process, RGB data are collected via DIC, while plasma stability is monitored using in-situ OES. These are paired with ellipsometer measurements of etch depth to train ANN and BNN models for predictive analysis.

## 3. Result and Discussion

### 3.1. Training set generation with ex-situ analysis

The results of the digital surface images are shown in **Figure 2**. The reference image, located at the top-right of Figure 2, corresponds to a 300 nm $SiO_2$ wafer. A distinct trend is observed in the etching behavior of $SiO_2$ under varying plasma conditions. As the top power increased from 50 to 110 W, the color of the etched samples become progressively lighter, indicating a higher etch rate. The effect of chamber pressure was also evident. At a pressure condition of 20

mTorr, samples exhibited more pronounced color changes compared to the other pressure conditions. In contrast, variations in the $CF_4$ flow rate showed minimal influence on surface color.

**Figure 3** presents the ellipsometry measurements of $SiO_2$ thickness after etching under different top power, $CF_4$ flow rate, and chamber pressure conditions. The measured thickness trends aligned well with the digital surface image results: more extensive etching occurred at higher power levels and lower pressure. The color gradient represents the remaining $SiO_2$ thickness, with lighter shades indicating greater etch depth. A detailed thickness scale is provided in the top-right corner for reference and is consistently applied across all 84 conditions. The size and measured thickness of the reference sample are indicated in the middle-right section.

A distinct trend is observed, where higher top power and lower chamber pressure conditions result in enhanced etching performance. As the top power increases from 50 to 110 W, the remaining $SiO_2$ thickness decreases consistently across all pressure conditions. Similarly, lower chamber pressure (20 mTorr) leads to more pronounced etching compared to higher pressures. Specifically, under conditions of 20 W bottom power and 110 W top power, the average thickness decreased from 302.98 nm to 193.23 nm, confirming that the etching depth exceeded 100 nm. This phenomenon can be attributed to the increased mean free path of ions at lower pressures, which enhances their kinetic energy and leads to a more efficient etching process.[36] However, variations in $CF_4$ flow rate (5-20 sccm) had minimal impact on the etching behavior. This negligible effect is attributed to the saturation of fluorine (F) radicals in the plasma, where an excess of $CF_4$ does not significantly increase the concentration active species participating in the etching reaction. Moreover, an excessive $CF_4$ flow may lead to the formation of carbon residues ($CF_X$), which can inhibit the etching process by depositing on the surface and reducing etch efficiency.[37, 38]

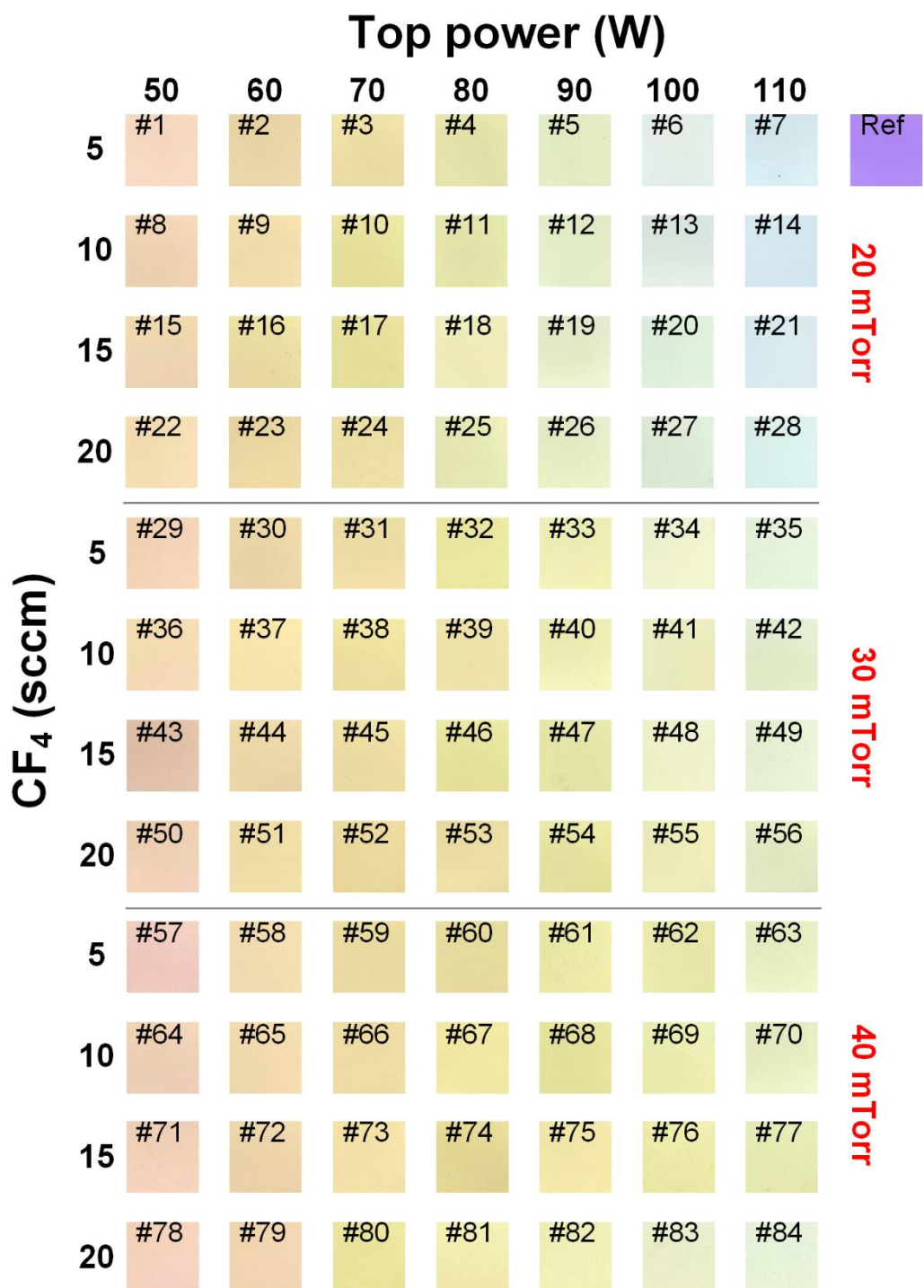

**Figure 2** Digital surface images of SiO$_2$ coupon wafer (1.5 × 1.5 cm) captured by iPhone 12. A total of 84 images, including a reference sample, were acquired under varying chamber pressures (20, 30, and 40 mTorr) and CF$_4$ flow rates (5, 10, 15, and 20 sccm). The results are segmented by horizontal lines to distinguish different pressure conditions.

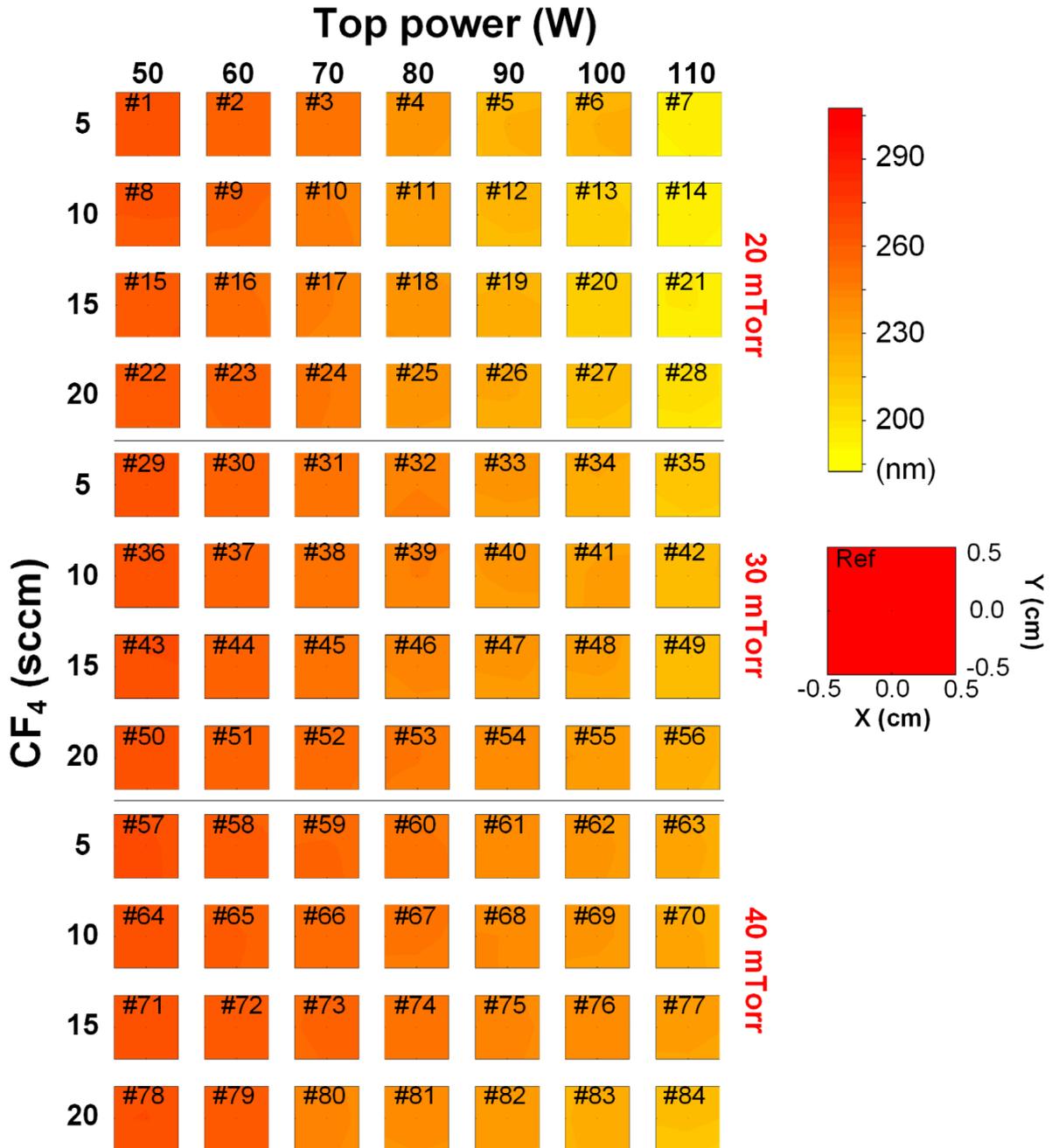

**Figure 3** The thickness contour of SiO$_2$ coupon wafer (1.5 × 1.5 cm) by ellipsometry mapping with 9 points. A total of 84 measurements, including a reference sample, were performed under varying chamber pressures (20, 30, and 40 mTorr) and CF$_4$ volume flow rates (5, 10, 15, and 20 sccm). Horizontal lines are used to separate results by pressure conditions.

### 3.2. In-situ plasma analysis

In our study, OES was employed to verify plasma stability and indirectly detect abnormal process conditions. The results of the OES are shown in **Figure 4**. Upon initiating plasma discharge with a three-gas mixture, a series of spectral peaks was observed. Among these, Ar, O$_2$, and CF$_3$ were identified as key species expected to form during the etching process. Consequently, Ar ($\lambda$ = 750 nm),[39] O$_2$ (630 nm),[40] and CF$_3$ (241.7 nm)[41] were monitored for analysis. All spectral data were normalized to compare analysis of intensity levels across different conditions. While minor fluctuations in peak intensities were intermittently observed, the overall emission patterns remained consistent. This consistency, though qualitative, supports the conclusion that the plasma was stable and the etching process proceeded under steady-state conditions.

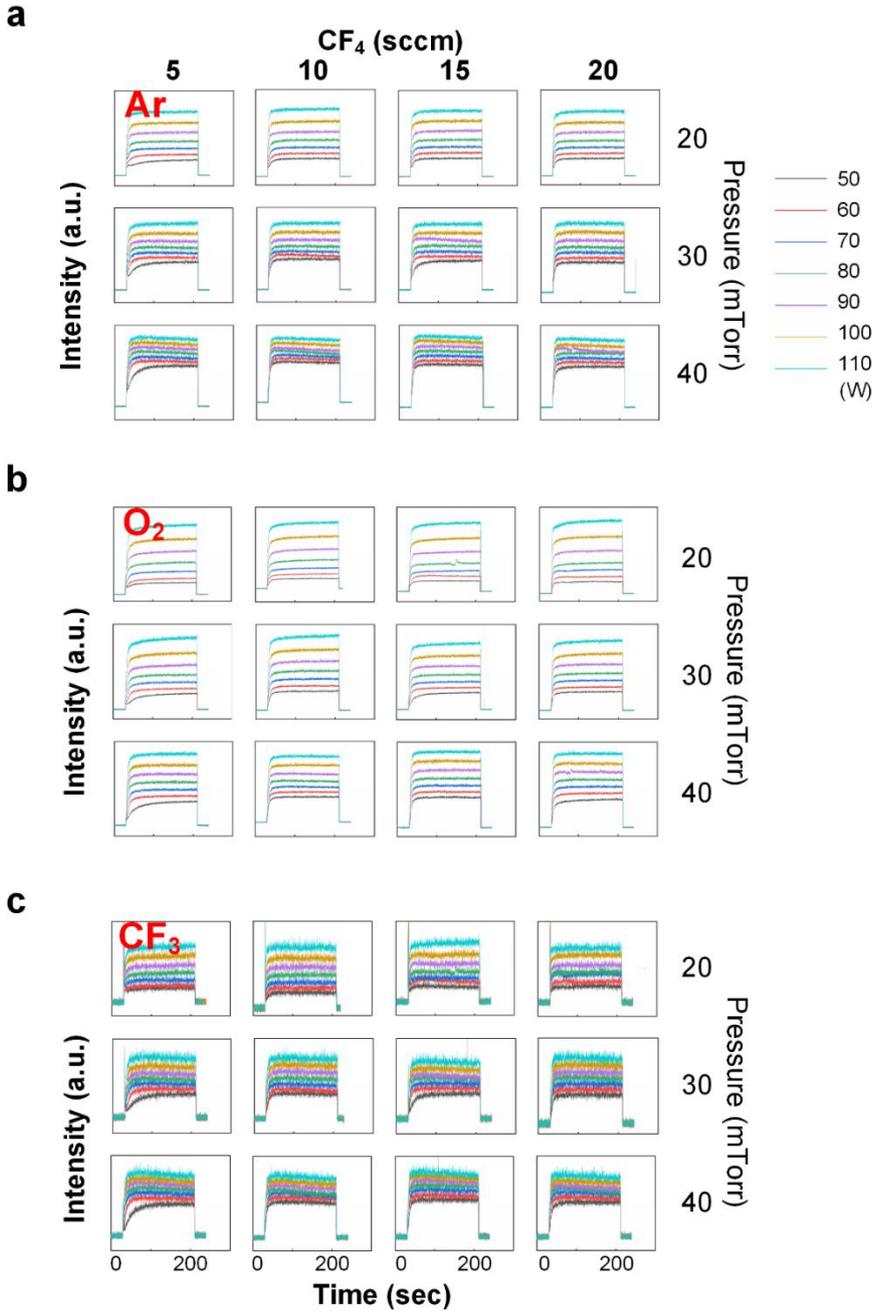

**Figure 4** Normalized optical emission spectra (a) Ar (b) $O_2$ and (c) $CF_3$, obtained under varying chamber pressures (20, 30, and 40 mTorr) and $CF_4$ flow rates (5, 10, 15, and 20 sccm). The emission profiles reflect stability of the plasma under process conditions, serving as indirect indicator of process steadiness.

**3.4. Etch depth training with ML models**

To establish the relationship between the process parameters, we consider neural network models. A deep neural network is a hierarchical model in which each layer applies a linear transformation followed by a nonlinearity to the preceding layer. In brief, we consider an ANN which takes an $n$-dimensional vector $x_0 \in R^n$ as an input, and outputs an $m$-dimensional vector $y \in R^m$. If the ANN consists of $k$ layers, the structure of the neural network is expressed as shown in **Eq. (1)**,

$$y(x_0; \{W^j\}, \{b^j\}) = \psi_k(W^k \psi_{k-1}(W^{k-1} \ldots \psi_2(W^2 \psi_1(W^1 x_0 + b^1) + b^2) \ldots + b^{k-1}) + b^k \tag{1}$$

where $W^j \in R^{d_j \times d_{j-1}}$ is a linear transformation $b^j \in R^{d_j}$ is a bias, and $\psi_j: R^{d_j} \to R^{d_j}$ is an element-wise nonlinear activation function acting on each component of $(W^j x_{j-1} + b^j)$. Given $N$ training data points $\{\bar{x}_i, \bar{y}_i: i = 1, \ldots, N\}$, a user can optimize the weights $W^j$ and biases $b^j$ by defining a loss function $L$. For example, a typical choice for $L$ is the Mean Squared Error (MSE) as defined in **Eq. (2)**.

$$L(W^1, \ldots, W^k, b^1, \ldots, b^k) = \frac{1}{N} \sum_{i=1}^{N} (y_i - \bar{y}_i)^2 \tag{2}$$

The problem of learning network weights is then formulated as the following optimization problem:

$$\{W^1, \ldots, W_k, b^1, \ldots, b^k\} = \operatorname{argmin} L(\{W^j\}, \{b^j\}) \tag{3}$$

A variety of computational techniques, including stochastic gradient descent and back-propagation, are available to solve the above problem. All neural network models used in this study are implemented using the PyTorch library.

**Table 2** presents example datasets collected from the experiment. The input variables consist of three parameters: pressure (p), volume flow rate of CF$_4$ ($\dot{Q}_{CF_4}$), and RF plasma power ($P$) of the reactor. The output is SiO$_2$ wafer thickness (T) measured using an ellipsometer at nine different points. Based on the initial thickness information, the etched thickness can be readily calculated. Using the available dataset, we evaluate the potential of ANN models in two distinct scenarios.

### 3.4.1. Prediction of etch depth from process parameters.

In the first scenario, we begin by predicting the average of nine measurements based on a given set of process parameters. In other words, $x_0 = \{p, \dot{Q}_{CF_4}, RF\}$. To this end, 7 samples are randomly selected from the total 84 as a validation dataset, leaving 77 (=84-7) samples for training. This validation dataset is summarized in **Table 3**. Although the amount of data is limited, we aim to explore the potential of using an ANN to predict etching performance. The neural network consists of an input layer with 3 features, a hidden layer with 32 neurons, and an output layer with a single neuron. The hidden layer applies a ReLU (Rectified Linear Unit) activation function, introducing nonlinearity to improve learning capabilities. Additionally, a dropout layer with a probability of 20% is included, which randomly deactivates some neurons during training to prevent overfitting. The model is trained using the Adam optimizer with weight decay for regularization.

Next, to support the use of the ANN approach and evaluate its potentials in a more qualitatively manner, we also construct a linear model,

$$y^l(x_0; W^l, b^l) = W^l x_0 \qquad (4)$$

and evaluate its error against the validation data. The result of validation test is summarized in **Table 3** and **Figure 5**. The MSE losses between the true values and the predictions from the ANN and linear models are 7.33 and 33.94, respectively. The lower MSE of the ANN model justifies the use of this approach, as it accounts for nonlinear relationships between the process parameters and thickness.

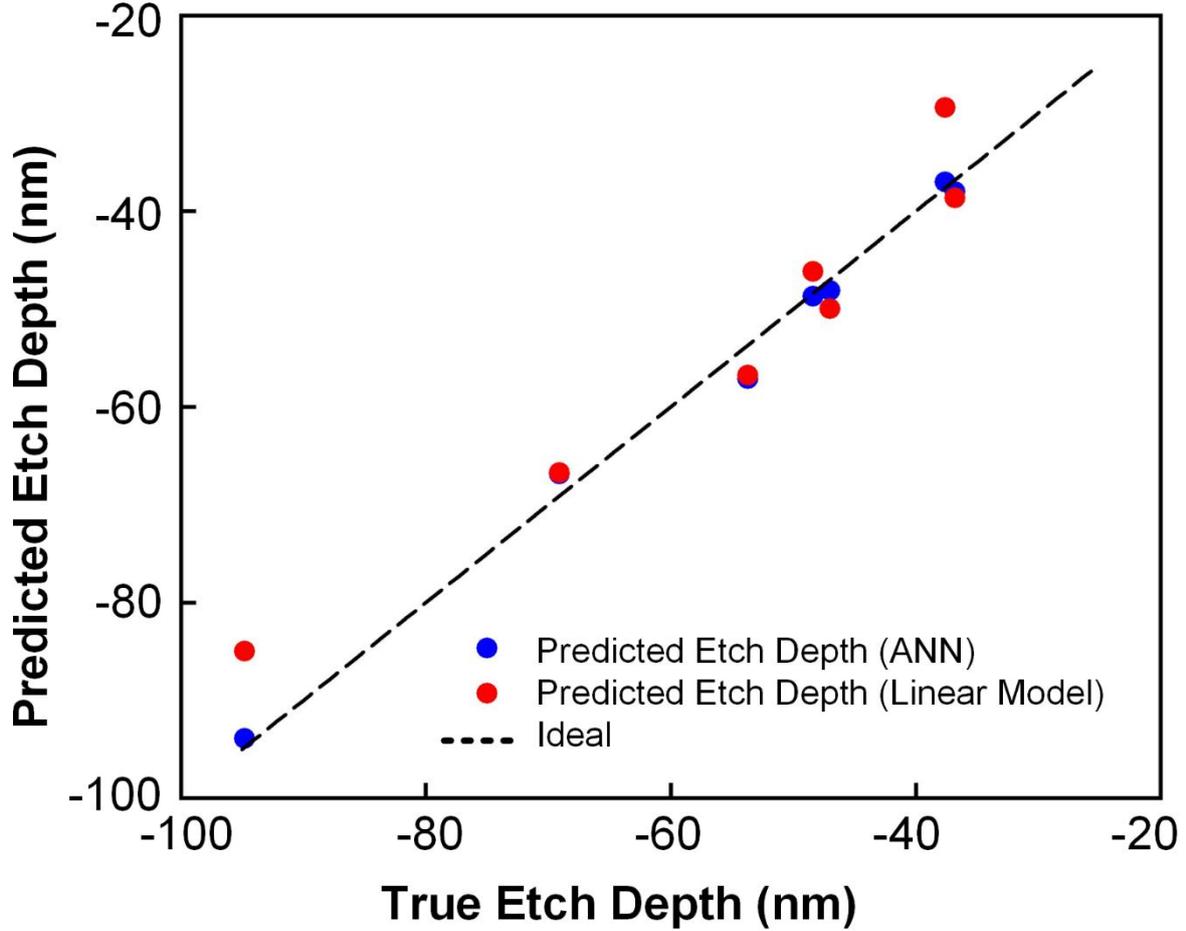

**Figure 5.** Etch depth predictions from the ANN model (blue), using process parameters {p, $\dot{Q}_{CF_4}$ P} compared to those from a linear model (orange) and the actual values (horizontal line).

**Table 2.** Example data set representing etched SiO$_2$ thickness.

| p [mTorr] | $\dot{Q}_{CF4}$ [sccm] | RF P [W] | T. #1 [nm] | T. #2 [nm] | T. #3 [nm] | T. #4 [nm] | T. #5 [nm] | T. #6 [nm] | T. #7 [nm] | T.#8 [nm] | T.#9 [nm] |
|---|---|---|---|---|---|---|---|---|---|---|---|
| 20 | 5 | 50 | 266.2 | 265.6 | 266.0 | 266.2 | 266.6 | 266.7 | 266.4 | 266.0 | 265.8 |
| 20 | 10 | 90 | 221.4 | 220.7 | 221.1 | 220.0 | 219.8 | 216.9 | 219.0 | 218.9 | 220.8 |
| 30 | 10 | 60 | 20 | 256.8 | 256.1 | 255.9 | 255.2 | 256.0 | 256.1 | 256.9 | 257.1 |
| 30 | 15 | 60 | 30 | 259.4 | 259.7 | 259.9 | 259.6 | 259.0 | 258.2 | 258.0 | 257.7 |
| 30 | 20 | 60 | 30 | 258.4 | 257.9 | 258.9 | 258.7 | 258.5 | 257.8 | 257.5 | 256.7 |

**Table 3.** Etch depth predictions from the ANN model (blue) from process parameters {p, $\dot{Q}_{CF_4}$, RF} compared to those from a linear model and the actual values. The MSE losses between the true values and the predictions from the ANN and linear models are 7.33 and 33.94, respectively.

| p | $\dot{Q}_{CF4}$ | RF P | True Etch Depth | Predicted Etch Depth (ANN) | Predicted Etch Depth (Linear model) |
|---|---|---|---|---|---|
| 40 | 15 | 80 | -53.70 | -57.13 | -56.76 |
| 20 | 5 | 50 | -36.80 | -38.00 | -38.65 |
| 40 | 5  | 70  | -48.40 | -48.70 | -46.17 |
| 20 | 20 | 60  | -47.00 | -48.10 | -49.98 |
| 20 | 10 | 100 | -94.80 | -93.90 | -84.97 |
| 40 | 15 | 50  | -37.60 | -37.02 | -29.41 |
| 20 | 10 | 80  | -69.10 | -66.84 | -66.73 |

Next, we extend the baseline predictive task by incorporating the variability present in the nine repeated measurements, with the goal of capturing both the uncertainty inherent in the data and the uncertainty in the model itself. Repeated measurements provide valuable information about data-level noise (aleatoric uncertainty), while a BNN can model uncertainty in the model parameters (epistemic uncertainty). By combining the two, one may obtain a more complete understanding of prediction reliability—especially important in semiconductor processes where both measurement error and model ambiguity are inevitable.

To retain the variability in repeated measurements, we treat each measurement as an individual data point rather than collapsing them into a mean value. This results in a dataset of 756 samples (9 measurements × 84 distinct process parameter combinations). Among these, 604 samples (approximately 80%) are used for training, and the remaining 152 for validation. To model uncertainty and quantify predictive confidence, we employ a BNN using the Monte Carlo (MC) Dropout technique. In this framework, model weights are implicitly treated as probabilistic by enabling dropout during both training and inference.[32] This allows the network to perform multiple stochastic forward passes for the same input, yielding a distribution of outputs that reflects model uncertainty. During inference, MC Dropout is performed by executing 50 stochastic forward passes for each validation input, thereby generating a distribution of predicted outputs. The mean of the sampled outputs is used as the final prediction, while the standard deviation represents the model's predictive uncertainty. To assess the model's reliability, the predicted uncertainty intervals are compared against the true etch depth values from the validation set.

A coverage analysis is performed to evaluate how often the true values fall within the predicted uncertainty intervals. Specifically, three categories are defined: (i) samples where the true value lies within the 1σ (68%) confidence interval, (ii) samples that lie outside the 1σ interval but within the 2σ (95%) interval, and (iii) samples that fall outside the 2σ interval. As visualized in **Figure 6**, 68.25% of the validation samples fall within the ±1σ range, while an additional 23.81% lie within the ±2σ range. Only 7.94% of the samples are outside the 2σ uncertainty bounds, indicating that the model successfully captures the variability of the data in most cases. These results indicate that the BNN effectively captures the underlying variability in the data and provides reliable uncertainty estimates. Such uncertainty quantification is especially valuable in semiconductor process modeling, where measurement noise and process fluctuations are inherent and critical to account for.

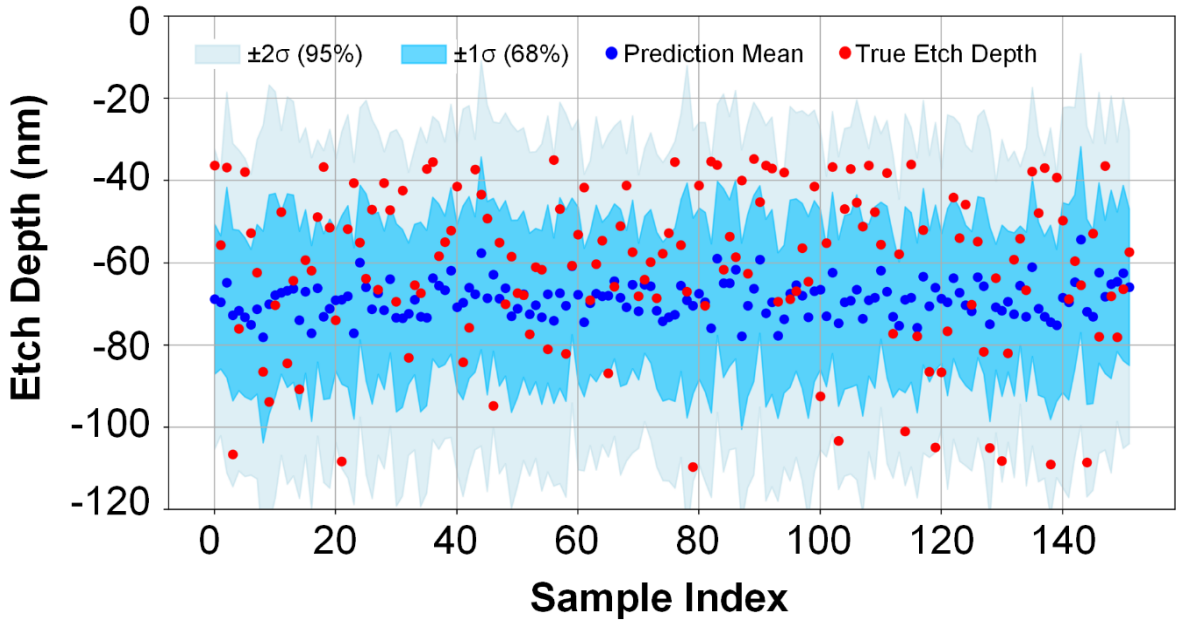

**Figure 6** Prediction of etch depth with confidence intervals using the Bayesian neural network model (blue dots) and actual etch depth values (red dots).

### 3.4.2. Prediction of etch depth from DIC

Next, we consider an ANN-based approach for predicting etching thickness from digital surface images. The underlying idea is that an ANN may be able to capture the relationship between the surface features shown in Figure 5 and the corresponding etching thickness illustrated in Figure 6. The primary motivation for exploring this relationship lies in the fact that acquiring digital surface images is significantly less costly than measuring actual thickness using an ellipsometer. Digital surface images were selected as a practical and cost-effective solution for real-time surface diagnostics. The use of digital imaging minimizes hardware modification and allows nonintrusive integration into existing process setups at minimal cost. Moreover, since this relationship is independent of specific plasma process parameters—such as $p$, $\dot{Q}_{CF_4}$, or RF power—the trained ANN model has the potential to generalize across different plasma processes.

Again, the number of ANN inputs is three, corresponding to the RGB channels of the digital surface images. Since the amount of training data is the same as in the previous case, we maintained the same neural network architecture. The model was trained using the Adam optimizer with weight decay applied for regularization. The results are summarized in **Table 4** and **Figure 8**, where the performance of the ANN is compared to that of a linear model. The MSE losses between the true values and the predictions from the ANN and linear models are 10.38 and 113.0, respectively. The substantially lower MSE of the ANN model compared to the linear model suggests that the relationship between RGB channels and etch thickness is more nonlinear than in previous cases, further justifying the use of the ANN-based approach.

**Table 4.** Etch depth predictions from the ANN model (blue) from RGB data of digital surface images, compared to those from a linear model (orange) and the actual values (horizontal line). The MSE validation loss between the true values and the predictions from the ANN and linear models are 10.38 and 113.0, respectively.

| R | G | B | True Etch Depth [nm] | ANN Model Prediction [nm] | Linear Model Prediction [nm] |
|---|---|---|---|---|---|
| 228.8 | 151.2 | 216.3 | -53.70 | -52.38 | -50.88 |
| 248.0 | 195.3 | 217.0 | -36.80 | -42.98 | -56.46 |
| 234.0 | 159.9 | 218.9 | -48.40 | -50.94 | -52.54 |
| 236.8 | 163.5 | 216.6 | -47.00 | -46.90 | -51.41 |
| 224.2 | 225.3 | 235.1 | -94.80 | -92.33 | -83.37 |
| 238.3 | 180.6 | 206.7 | -37.60 | -37.18 | -51.32 |
| 227.2 | 170.1 | 228.4 | -69.10 | -64.62 | -62.69 |

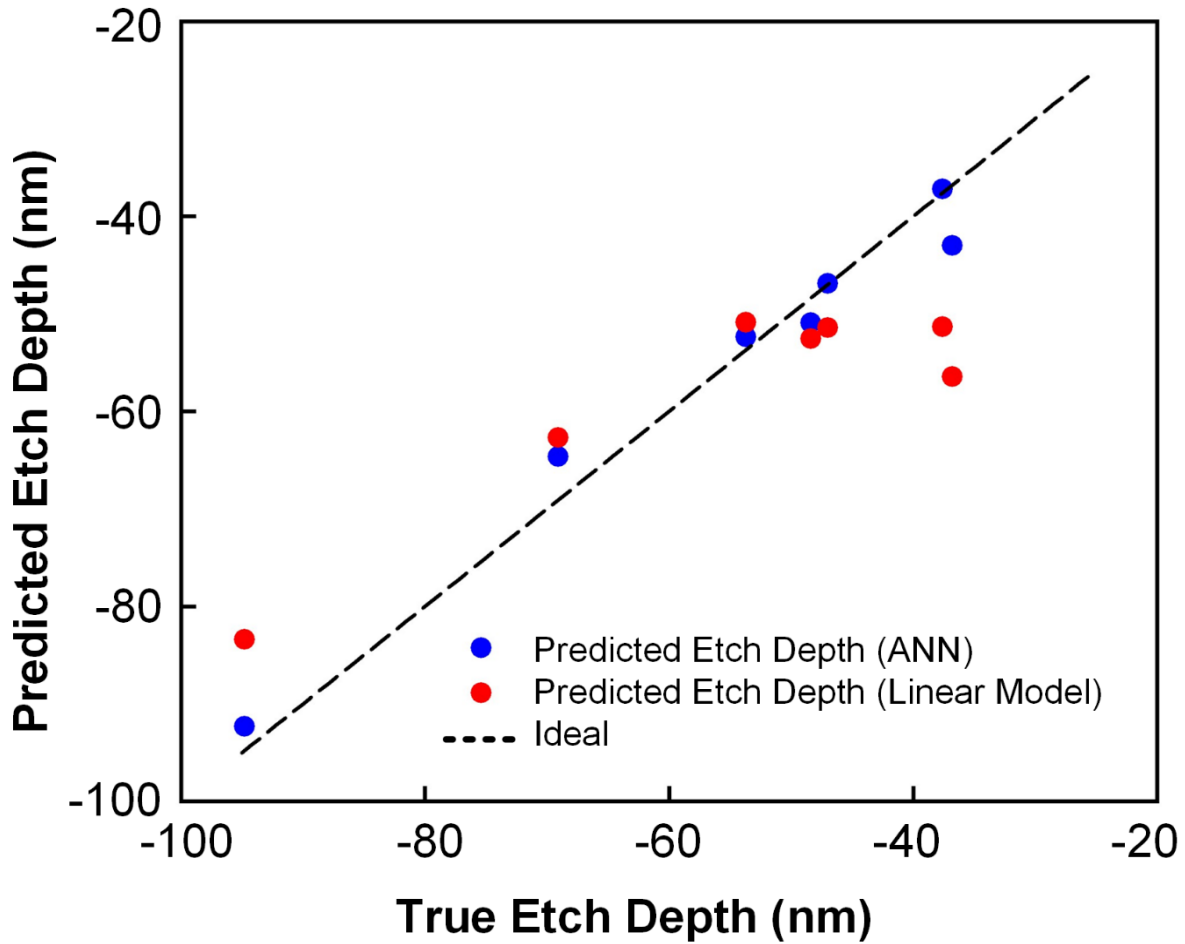

**Figure 7** Etch depth predictions from the ANN model (blue) using RGB data of digital surface images compared to those from a linear model (red) and the actual values (horizontal line).

Finally, we applied the same BNN approach using RGB digital surface inputs as model features. Each repeated RGB measurement was treated as an independent data point to preserve data-level variability, and MC Dropout was employed to estimate predictive uncertainty through multiple stochastic forward passes.

A coverage analysis was conducted to assess the reliability of the model's uncertainty estimates. As illustrated in Figure 8, 63.16% of the validation samples fell within the ±1σ (68%) confidence interval, and an additional 34.87% were within the ±2σ (95%) range. Only 1.97% of the samples lay outside the ±2σ bounds, indicating that the model successfully captured the variability present in the RGB data. These results confirm that the BNN framework provides robust uncertainty quantification even when using high-dimensional digital image features—an important capability for modeling semiconductor processes where both measurement noise and process variation are unavoidable.

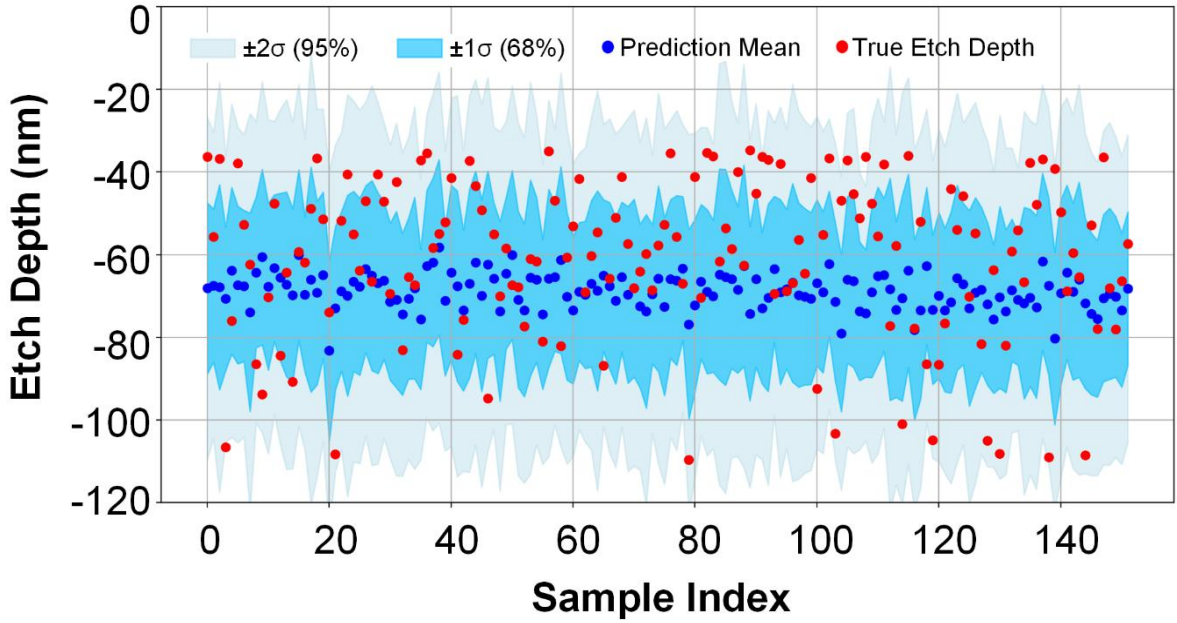

**Figure 8** DIC prediction of etch depth and confidence intervals using the Bayesian neural network model (blue dots) and actual etch depth values (red dots).

## 4. Conclusion

This study presents a non-contact, machine learning (ML)-based approach for in-situ monitoring of etch depth in semiconductor manufacturing, offering a cost-effective and real-time alternative to conventional ex-situ techniques. Two complementary scenarios were explored. In the first, etch depth was predicted from process parameters using both ANN and BNN. The ANN effectively captured nonlinear relationships, while the BNN enabled uncertainty quantification, providing insights into the reliability of the predictions. In the second scenario, RGB data from digital surface images were used as inputs, demonstrating that meaningful predictions can be achieved even in the absence of explicit process parameters. The BNN effectively captured both aleatoric and epistemic uncertainties.

The integration of DIC with ML opens new possibilities for real-time diagnostics in plasma etching processes. By eliminating the need for chamber interruption and reducing dependence on expensive instrumentation, the proposed framework enhances process stability and has the potential to improve device yield. Furthermore, the demonstrated capacity of BNN to quantify uncertainty provides critical reliability insights for high-precision manufacturing environments. Future work will aim to expand the dataset and explore additional image features to further improve model robustness and generalizability across different etching conditions.

**Acknowledgements**

This work was supported by the Hongik University new faculty research support fund.

**References**

1. Romero, M.; Sanz, M.; Tanarro, I.; Jiménez, A.; Muñoz, E., *J. Phys. D: Appl. Phys.* **2010,** *43* (49), 495202.
2. Radamson, H. H.; Miao, Y.; Zhou, Z.; Wu, Z.; Kong, Z.; Gao, J.; Yang, H.; Ren, Y.; Zhang, Y.; Shi, J., *Nanomater.* **2024,** *14* (10), 837.
3. Ji, Y. J.; Kim, K. S.; Kim, K. H.; Byun, J. Y.; Yeom, G. Y., *Appl. Sci. Converg. Technol* **2019,** *28* (5), 142-147.
4. Tsai, S.-J.; Wang, C.-L.; Lee, H.-C.; Lin, C.-Y.; Chen, J.-W.; Shiu, H.-W.; Chang, L.-Y.; Hsueh, H.-T.; Chen, H.-Y.; Tsai, J.-Y., *Sci. Rep.* **2016,** *6* (1), 28326.
5. Geng, K.; Chen, D.; Zhou, Q.; Wang, H., *Electronics* **2018,** *7* (12), 416.
6. Das, S.; Sebastian, A.; Pop, E.; McClellan, C. J.; Franklin, A. D.; Grasser, T.; Knobloch, T.; Illarionov, Y.; Penumatcha, A. V.; Appenzeller, J., *Nat. Electron.* **2021,** *4* (11), 786-799.
7. Yan, S.; Cong, Z.; Lu, N.; Yue, J.; Luo, Q., *Sci. China Inf. Sci.* **2023,** *66* (10), 200404.
8. Lim, N.; Efremov, A.; Kwon, K.-H., *Thin Solid Films* **2019,** *685*, 97-107.
9. Gasvoda, R. J.; Zhang, Z.; Wang, S.; Hudson, E. A.; Agarwal, S., *J. Vac. Sci. Technol. A* **2020,** *38* (5).
10. Huff, M., *Micromachines* **2021,** *12* (8), 991.
11. Pasquariello, D.; Hedlund, C.; Hjort, K., *J.Electrochem. Soc.* **2000,** *147* (7), 2699.
12. Kim, C.; Lee, S.; Kim, M.; Choi, M. S.; Kim, T.; Kim, H.-U., *Appl. Sci. Converg. Technol.* **2023,** *32* (5), 106-109.
13. Kim, C.; Kim, M.; Cho, D.; Kanade, C. K.; Seok, H.; Bak, M. S.; Kim, D.; Kang, W. S.; Kim, T.; Kim, H.-U., *Chem. Mater.* **2023,** *35* (3), 1016-1028.
14. Dragoi, V.; Lindner, P., *ECS transactions* **2006,** *3* (6), 147.
15. King, S. W.; Smith, L. L.; Barnak, J. P.; Ku, J.-H.; Christman, J. A.; Benjamin, M. C.; Bremser, M. D.; Nemanich, R. J.; Davis, R. F., *MRS Online Proceedings Library (OPL)* **1995,** *395*, 739.
16. Fan, Y.; Li, J.; Guo, Y.; Xie, L.; Zhang, G., *Measurement* **2021,** *171*, 108829.
17. Pounds, K.; Bao, H.; Luo, Y.; De, J.; Schneider, K.; Correll, M.; Tong, Z., *ACS Food Sci. Technol.* **2022,** *2* (7), 1123-1134.
18. Granados-Vega, B. V.; Maldonado-Flores, C.; Gómez-Navarro, C. S.; Warren-Vega, W. M.; Campos-Rodríguez, A.; Romero-Cano, L. A., *Foods* **2023,** *12* (20), 3829.
19. Masawat, P.; Harfield, A.; Namwong, A., *Food Chem.* **2015,** *184*, 23-29.
20. Huang, J.; Sun, J.; Warden, A. R.; Ding, X., *Food Control* **2020,** *108*, 106885.
21. de Oliveira Krambeck Franco, M.; Suarez, W. T.; Santos, V. B. d., *Food Anal. Methods* **2017,** *10*, 508-515.
22. Permana, M. D.; Sakti, L. K.; Luthfiah, A.; Firdaus, M. L.; Takei, T.; Eddy, D. R.; Rahayu, I., *Trends Sci.* **2023,** *20* (4), 5149-5149.
23. Bang-iam, N.; Udnan, Y.; Masawat, P., *Microchem. J.* **2013,** *106*, 270-275.
24. Aydın, Ö. F.; Aydın, M.; Demir, M. C.; Kahraman, S., *Chemometrics Intellig. Lab. Syst.* **2025,** *257*, 105310.
25. Firdaus, M. L.; Alwi, W.; Trinoveldi, F.; Rahayu, I.; Rahmidar, L.; Warsito, K., *Procedia Environ. Sci.* **2014,** *20*, 298-304.
26. Bhatt, S.; Kumar, S.; Gupta, M. K.; Datta, S. K.; Dubey, S. K., *Meas. Sci. Technol.* **2023,** *35* (1), 015030.
27. Choi, W.; Shin, J.; Hyun, K.-A.; Song, J.; Jung, H.-I., *Biosens. Bioelectron.* **2019,** *130*, 414-419.

28. Lee, S.; Park, Y.; Liu, P.; Kim, M.; Kim, H.-U.; Kim, T., *Sensors* **2023,** *23* (19), 8226.
29. Kang, M.; Jang, S. K.; Kim, J.; Kim, S.; Kim, C.; Lee, H.-C.; Kang, W.; Choi, M. S.; Kim, H.; Kim, H.-U., *J. Sens. Actuator Netw.* **2024,** *13* (6), 75.
30. Park, Y.; Liu, P.; Lee, S.; Cho, J.; Joo, E.; Kim, H.-U.; Kim, T., *Sensors* **2023,** *23* (12), 5563.
31. Chernatynskiy, A.; Phillpot, S. R.; LeSar, R., *Annu. Rev. Mater. Res.* **2013,** *43* (1), 157-182.
32. Gal, Y.; Ghahramani, Z. In *Dropout as a bayesian approximation: Representing model uncertainty in deep learning*, international conference on machine learning, PMLR: **2016**; pp 1050-1059.
33. Xue, Y.; Cheng, S.; Li, Y.; Tian, L., *Optica* **2019,** *6* (5), 618-629.
34. Kolpaková, A.; Kudrna, P.; Tichý, M. In *Study of plasma system by OES (optical emission spectroscopy)*, Proc. of 20th Annual Conference of Doctoral Students, **2011**; pp 180-185.
35. Charles, B.; Fredeen, K. J., *Perkin Elmer Corp* **1997,** *3* (2), 115.
36. Lieberman, M. A.; Lichtenberg, A. J., *MRS Bulletin* **1994,** *30* (12), 899-901.
37. Zhao, S.-X.; Gao, F.; Wang, Y.-N.; Bogaerts, A., *Plasma Sources Sci. Technol.* **2012,** *22* (1), 015017.
38. Jenq, J.-S.; Ding, J.; Taylor, J. W.; Hershkowitz, N., *Plasma Sources Sci. Technol.* **1994,** *3* (2), 154.
39. Sainct, F. P.; Durocher-Jean, A.; Gangwar, R. K.; Mendoza Gonzalez, N. Y.; Coulombe, S.; Stafford, L., *plasma* **2020,** *3* (2), 38-53.
40. Xu, W.; Mao, X.; Zhou, N.; Zhang, Q.-Y.; Peng, B.; Shen, Y., *Vacuum* **2022,** *196*, 110785.
41. Suto, M.; Washida, N., *J. Chem. Phys.* **1983,** *78* (3), 1012-1018.